\pdfoutput=1

\documentclass[11pt]{article}

\usepackage{EMNLP2023}

\usepackage{times}
\usepackage{latexsym}
\usepackage{booktabs}
\usepackage{graphicx}
\usepackage{enumitem}
\usepackage{amsmath}
\usepackage{adjustbox}

\usepackage[T1]{fontenc}

\usepackage[utf8]{inputenc}

\usepackage{microtype}

\usepackage{inconsolata}

\title{ExpNote: Black-box Large Language Models are Better Task Solvers \\ with Experience Notebook}

\author{
  Wangtao Sun$^{1,2}$, Xuanqing Yu$^{2,3}$, Shizhu He$^{1,2}$, Jun Zhao$^{1,2}$, Kang Liu$^{1,2,4}$ \\
  \textit{$^{1}$The Laboratory of Cognition and Decision Intelligence for Complex Systems,} \\
  \textit{Institute of Automation, Chinese Academy of Sciences, Beijing, China} \\
  \textit{$^{2}$School of Artificial Intelligence, University of Chinese Academy of Sciences, Beijing, China} \\
  \textit{$^{3}$CAS Engineering Laboratory for Intelligent Industrial Vision,} \\
  \textit{Institute of Automation, Chinese Academy of Sciences, Beijing, China} \\
  \textit{$^{4}$Shanghai Artificial Intelligence Laboratory} \\
  \texttt{\{sunwangtao2021, yuxuanqing2021\}@ia.ac.cn} \\
  \texttt{\{shizhu.he, jzhao, kliu\}@nlpr.ia.ac.cn}
}

\begin{document}
\maketitle

\begin{abstract}
Black-box Large Language Models (LLMs) have shown great power in solving various tasks and are considered general problem solvers. However, LLMs still fail in many specific tasks although understand the task instruction. In this paper, we focus on the problem of boosting the ability of black-box LLMs to solve downstream tasks. We propose ExpNote, an automated framework to help LLMs better adapt to unfamiliar tasks through reflecting and noting experiences from training data and retrieving them from external memory during testing. We evaluate ExpNote on multiple tasks and the experimental results demonstrate that the proposed method significantly improves the performance of black-box LLMs. The data and code are available at https://github.com/forangel2014/ExpNote.
\end{abstract}

\section{Introduction}
\noindent
Large Language Models (LLMs) have demonstrated astonishing capabilities in natural language understanding and generation \cite{cot, huang2022language, paradigm, bang2023multitask, chen2024comm}. However, due to the limited parameters and context processing length, LLMs are not able to master all task-specific knowledge in real-world applications. As a result, LLMs may perform mediocre on some specific tasks, such as inductive reasoning \cite{bang2023multitask} and entity recognition \cite{chen2023large}. 

Therefore, how to make LLMs adapt to the downstream tasks has tracked more and more attention. Recent techniques such as prefix-tuning \cite{prefix-tuning}, P-tuning \cite{p-tuning} and LoRA \cite{lora} proposed low-cost solutions for fine-tuning LLMs. However, these methods are not capable of black-box powerful LLMs, such as ChatGPT and GPT4 \cite{openai2023gpt4}.

\begin{figure}[t]
    \centering
    \begin{adjustbox}{max width=\columnwidth}    
    \includegraphics[width=1.0\textwidth]{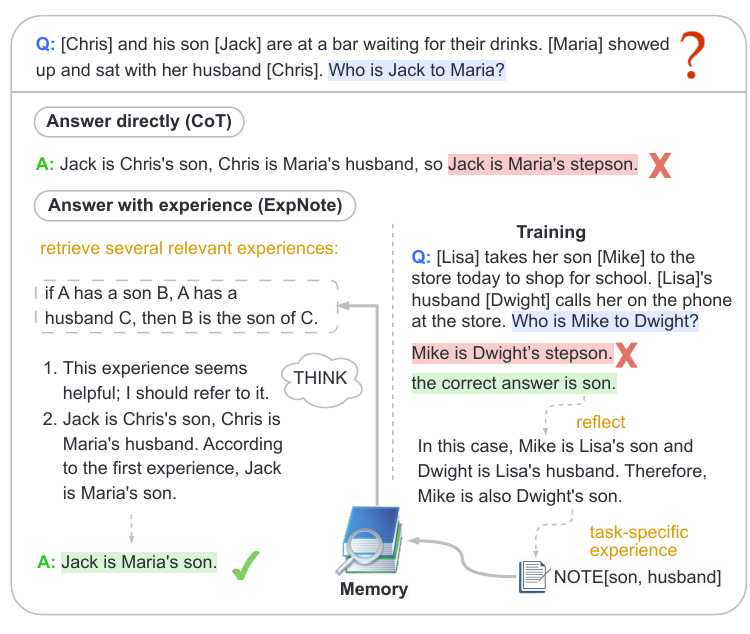}
    \end{adjustbox}
    \caption{An illustration of how ExpNote assists LLM in enhancing the effectiveness of task-solving. ExpNote can automatically generalize relevant experiences from other samples and apply them to specific tasks.}
    \label{figure1}
\end{figure}

To empower black-box LLMs on specific tasks, several works \cite{memprompt, teachme} have focused on equipping the LLMs with external dynamic memory to store useful task-specific knowledge and facts. However, these task-specific knowledge and facts in the memory usually come from expert annotation or human feedback, which is very costly to obtain. 
On the other hand, several researchers \cite{self-reflection, self-refine, rl4f} try to exploit the reflection ability of LLMs to automatically generate such knowledge for specific tasks. 
However, most reflection-based methods are only able to empower the LLMs in the same case, without the ability to generalize to other instances.

Thus, this paper proposes a framework ExpNote (Experience Notebook), to empower black-box LLMs on downstream tasks by learning and using task-specific experience automatically. We equip the LLMs with a dynamic memory and design several commands to help LLMs interact with it. 
In specific, in the training stage, ExpNote guides LLMs to generate task-specific experiences and store them in an external memory. In the testing stage, ExpNote uses a retriever to retrieve relevant experiences from the memory, the learned experiences will help the LLMs to solve the cases which they failed to answer directly (Figure~\ref{figure1}). 

We evaluate ExpNote on multiple tasks. The results show that ExpNote can empower the LLMs effectively, and significantly outperform other prompting methods like ordinary in-context learning (CoT, \citealt{cot}), memory-based method (TeachMe, \citealt{teachme}), and case-by-case reflection-based method (Reflexion, \citealt{self-reflection}).

Moreover, we empirically compared different types of experiences to examine their effectiveness in helping LLMs adapt to unfamiliar tasks. Specifically, we compared learned task-specific experiences with original cases and experiences learned from positive cases (succeeded) and negative cases (failed). We find that prompting with experiences is more helpful than original cases for LLMs to generalize to new cases, and experiences from both positive and negative cases are beneficial.

The major contributions of this paper are two-fold:
\begin{itemize}[itemsep=1pt,topsep=1pt,parsep=0pt,leftmargin=*]
    \item{We propose a framework ExpNote to empower the LLMs in various tasks through interacting with dynamic memory. ExpNote conducts fully automated reflection, noting, and retrieval, without the need of any annotated knowledge and facts or any human feedback.}
    
    \item{We investigate different types of experiences and show the learned task-specific experiences help LLMs to better generalize than the original cases in the task, and experiences from both positive and negative cases are beneficial.}
\end{itemize}

\begin{figure*}[t]
    \centering
    \begin{adjustbox}{max width=\textwidth}    
    \includegraphics[width=1.0\textwidth]{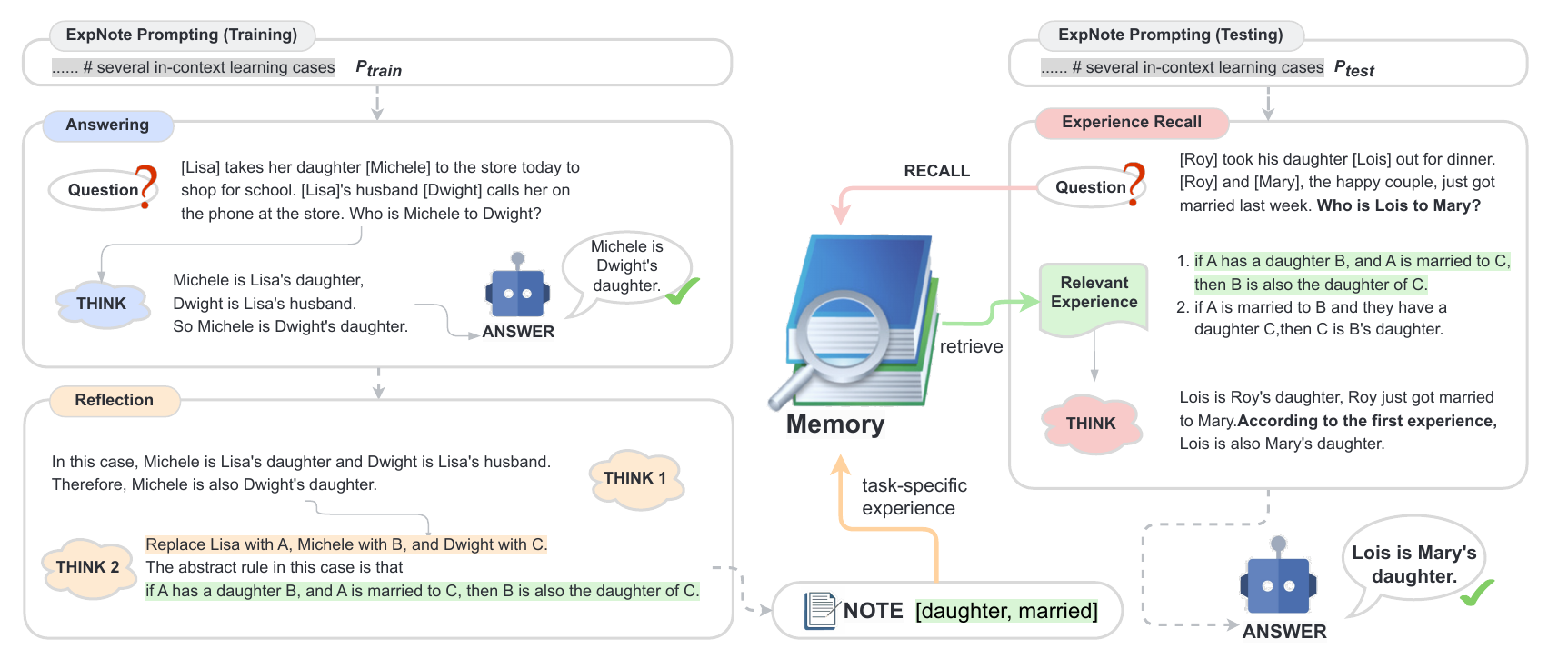}
    \end{adjustbox}
    \caption{The framework of ExpNote. This framework shows how LLMs use ExpNote to solve specific tasks, including the training (left) and testing (right) stages.}
    \label{figure2}
\end{figure*}

\section{Related Work}

\subsection{Language Model Taking Action}
\noindent
In order to address the lack of knowledge, reasoning, and specific abilities in large models, many efforts have focused on utilizing external knowledge sources and tools to assist large models in completing tasks. Toolformer \cite{toolformer} proposed fine-tuning on API-augmented datasets to enable the LLMs to master the ability to use external tools with API calls, thereby improving the performance of the LLMs in a series of tasks. ReAct \cite{react} proposed that by using a Wikipedia search API and generating trajectories similar to human thinking, the LLMs can utilize external knowledge during reasoning and provide interpretable reasoning paths. HuggingGPT \cite{hugginggpt} proposes to solve any AI task by using the models on the huggingface as its toolkit. 

\subsection{Language Model with Dynamic Memory}
\noindent
Some existing works have noticed the need to equip LLMs with dynamic memory. MemPrompt \cite{memprompt} retrieves the stored user feedback of the intention for similar questions to enhance the current prompt for the LLMs. TeachMe \cite{teachme} allows LLMs to store the missing and wrong facts during the QA task with the correction of user feedback. These methods created a new paradigm to boost the ability of LLMs in a general way. However, they rely heavily on human feedback or annotated facts. 
REMEMBERER \cite{remember} proposed to consider LLMs as a semi-parametric RL agent. It trains LLMs to take the next action based on the retrieved (observation, action, Q-value) tuple.

\subsection{Language Model Reflection}
\noindent
Recently, some works have been proposed to correct the mistakes of LLMs in conducting specific tasks by using their capacity of self-reflection. Reflexion \cite{self-reflection} focused on sequential decision-making tasks. A heuristic function is adopted to judge whether the trial is successful or not. And LLMs will reflect those trials that are thought to have failed. The reflective information will be used to support the LLMs in improving their own decision-making process in the next trial. Self-refine \cite{self-refine} proposed a method to iteratively improve the output of a large model through its own feedback, achieving improvements in multiple generation tasks. 
However, these reflection methods are limited to certain cases, without being abstract and able to generalize to other data points.

\subsection{Language Model Thought Chain}
Furthermore, there have been some efforts to improve the reasoning performance of LLMs by enhancing their thought chains in specific tasks. For example, DIVERSE \cite{li2023making}  proposed a method that generates multiple different reasoning paths and uses a validator for weighted voting to filter out incorrect answers. However, this method demands manual construction of reasoning paths for each training question-answer pair and extensive human evaluation, restricting its use on large datasets.

Drozdov \cite{drozdov2022compositional} and others introduced a technique that decomposes complex questions into sub-questions and provides answers to these sub-questions, serving as hints to assist the model in reaching the final answer. Faithful CoT \cite{lyu2023faithful} , on the other hand, prompts a language model to translate complex queries into a reasoning chain that includes question decomposition and corresponding symbolic language solving, thus enhancing interpretability. 

These approaches offer intriguing ideas for improving the reasoning performance of LLMs but still face challenges related to the need for substantial high-quality annotations, difficulties in reusing experiences, and sample generalization.

\section{ExpNote}
\subsection{The Framework}
\noindent
As shown in Figure~\ref{figure2}, all the tasks are formalized as the tuple $(x, y)$, where $x$ is the input question and $y$ is the desired answer. For each task, we write prompts $P_{train}$ and $P_{test}$ that encourage the LLM to use ExpNote following the illustrations. In the training stage, LLM is first ordered to infer the answer like ordinary CoT reasoning. 
\begin{equation}
    \hat{y} \sim p_{LLM} (\cdot|P_{train}, x)
\end{equation}
\noindent 
After the answer is obtained, the ExpNote will produce feedback $F(\hat{y}, y)$ to LLM depending on whether $\hat{y} = y$. Note that this feedback $F(\hat{y}, y)$ only includes a simple prompt containing the ground-truth of the current question, without any additional knowledge like TeachMe \cite{teachme}. Then LLM is supposed to reflect 
and store the learned experience $e$ into the memory. 

\begin{equation}
    e \sim p_{LLM} (\cdot|P_{train}, x, F(\hat{y}, y))
\end{equation}
\noindent
Where $e$ is a key-value pair of learned task-specific experience, e.g. key = \emph{daughter, married} and value = \emph{if A has a daughter B, and A is married to C, then B is also the daughter of C} (Figure~\ref{figure2}).
This process is achieved by taking $n$ extra actions to interact with the memory, which we will describe in Sec~\ref{command}. 

In the testing stage, ExpNote will use the testing instance as the search query to retrieve $k$ experiences from the dynamic memory. The retrieved experiences will be added to the prompts for the LLM. Then LLM will decide whether to refer to these experiences and finally output the answer.

\begin{equation}
    \begin{aligned}
       & \{e_i\}_{i=1}^{k} = Retrieve(x, k) \\
       & \hat{y} \sim p_{LLM} (\cdot|P_{test}, x, \{e_i\}_{i=1}^{k})
    \end{aligned}
\end{equation}

The full examples of ExpNote on different tasks are shown in Appendix~\ref{example}.

\subsection{Interaction Commands} \label{command}
ExpNote designs several commands for LLMs to interact with the memory, including summarizing and applying the experiences (\textbf{THINK}), storing the experiences in the memory (\textbf{NOTE}), and retrieving relevant experiences for the testing instances (\textbf{RECALL}). They are described in detail as follows:

\begin{itemize}[itemsep=1pt,topsep=1pt,parsep=0pt,leftmargin=*]
    \item{\textbf{THINK[arg]}. Inspired by ReAct \cite{react}, in both training and testing stages, we enable LLM to use command \textbf{THINK} to organize its current thoughts and make the next decision.}
    \item{\textbf{NOTE[arg1]: arg2}. In the training stage, we prompt the LLMs to use this command \textbf{NOTE} after answering each question. The command \textbf{NOTE} will store the experience \textbf{arg2} as a value in the memory with its key \textbf{arg1}.}
    \item{\textbf{RECALL[arg]}. In the testing stage, this command is automatically executed by ExpNote at the beginning of each question to recall relevant experiences. ExpNote will use a retriever to retrieve up to $k$ relevant experiences using the question \textbf{arg} as the search query. The content of these experiences will then be added to the prompt.}
\end{itemize}

\section{Experiments}
In this section, we want to answer the following research questions:
\begin{itemize}[itemsep=1pt,topsep=1pt,parsep=0pt,leftmargin=*]
    \item{\textbf{RQ1}. Is ExpNote able to help LLMs to adapt to new tasks effectively?}
    \item{\textbf{RQ2}. Which kinds of experiences help LLMs solve tasks better?}
\end{itemize}

\subsection{Datasets}
\noindent
To show the effectiveness of ExpNote in handling various tasks, we select multiple datasets from different tasks for empirical study, including CLUTRR (inductive reasoning, \citealt{clutrr}), METS-CoV (medical entity recognition, \citealt{mets}), EMOJI (text-to-emoji prediction, \citealt{emoji}). Besides, we propose a dataset LETS (letter splicing) to evaluate the symbolic reasoning capability of LLM enhanced with ExpNote. The detail of LETS can be found in Appendix~\ref{lets}. For each task, we tested ExpNote and other methods on 100 cases. The other setups of the experiments can be found in Appendix~\ref{setup}.

\subsection{Baselines}
\noindent
To answer the \textbf{RQ1}, apart from basic zero-shot and few-shot settings, we select the ordinary in-context learning method (CoT, \citealt{cot}), memory-based method (TeachMe, \citealt{teachme}) and case-by-case reflection-based method (Reflexion, \citealt{self-reflection}) for comparison.
\begin{itemize}[itemsep=1pt,topsep=1pt,parsep=0pt,leftmargin=*]
    \item{CoT \cite{cot}: Several cases of solving the task using Chain-of-Thought are shown to the LLM.}
    \item{TeachMe \cite{teachme}: As the core facts or human feedback of these specific tasks are hard to obtain, we adopt a commonsense knowledge base, Conceptnet \cite{speer2017conceptnet}, to serve as the memory for TeachMe.}
    \item{Reflexion \cite{self-reflection}: As the heuristic function of repetitive action detection described in Reflexion is not working for these tasks. Thus we allow Reflexion to do a little cheating: it is allowed to try again after every failed trial without being informed of the ground-truths. This setting is equivalent to obtaining a golden function that accurately determines the success/failure of each trial with a 100\% success rate.}
\end{itemize}

To answer the \textbf{RQ2}, we have also implemented several variants of ExpNote:
\begin{itemize}[itemsep=1pt,topsep=1pt,parsep=0pt,leftmargin=*]
    \item{\emph{disabled}: This variant adopts the reasoning form of ExpNote while disabling its retrieval function.}
    \item{\emph{case}: This variant will recall the original questions and answers of the noted cases instead of the learned experiences.}
    \item{\emph{positive} / \emph{negative}: This variant only retains the experiences that learned from the training sample which LLMs answered correctly / incorrectly.}
\end{itemize}

\begin{table}[t]
\centering
\begin{adjustbox}{width=\columnwidth}
\begin{tabular}{lccccl}
    \toprule
    &methods & CLUTRR & METS & EMOJI & LETS \\
    \midrule
    &zero-shot  & 36 & 23 & 34 & 35 \\
    &few-shot  & 44 & 61 & 47 & ~~1 \\
    &CoT & 40 & 54 & 54 & 60  \\
    &TeachMe & 31 & 51 & 56 & 56  \\ 
    &Reflexion & 54 & 62 & 71 & 68 \\
    \midrule
    &ExpNote & \textbf{61} & \textbf{66} & \textbf{74} & \textbf{89}\\
    \bottomrule
\end{tabular}
\end{adjustbox}
\caption{Accuracy of ExpNote and baselines on 4 datasets.}
\label{table1}
\end{table}

\begin{table}[t]
\centering
\begin{adjustbox}{width=\columnwidth}
\begin{tabular}{lcllll}
    \toprule
    &variants & CLUTRR & METS & EMOJI & LETS \\
    \midrule
    &\emph{disabled} & 35 (0) & 49 (0) & 57 (0) & 50 (0) \\
    &\emph{case} & 49 (128) & 59 (279) & 58 (20) & 51 (87)\\
    &\emph{positive} & 51 (73) & 56 (166) & 66 (14) & 64 (41)\\
    &\emph{negative} & 55 (55) & 52 (113) & 60 (6)& 71 (46)\\
    \midrule
    &ExpNote & \textbf{61} (128) & \textbf{66} (279) & \textbf{74} (20) & \textbf{89} (87)\\
    \bottomrule
\end{tabular}
\end{adjustbox}
\caption{Accuracy of ExpNote and its variants on 4 datasets. The numbers in the small bracket are numbers of experiences stored in the memory.}
\label{table2}
\end{table}

\begin{figure}[t]
    \centering
    \begin{adjustbox}{max width=\columnwidth}
    \includegraphics[width=0.75\textwidth]{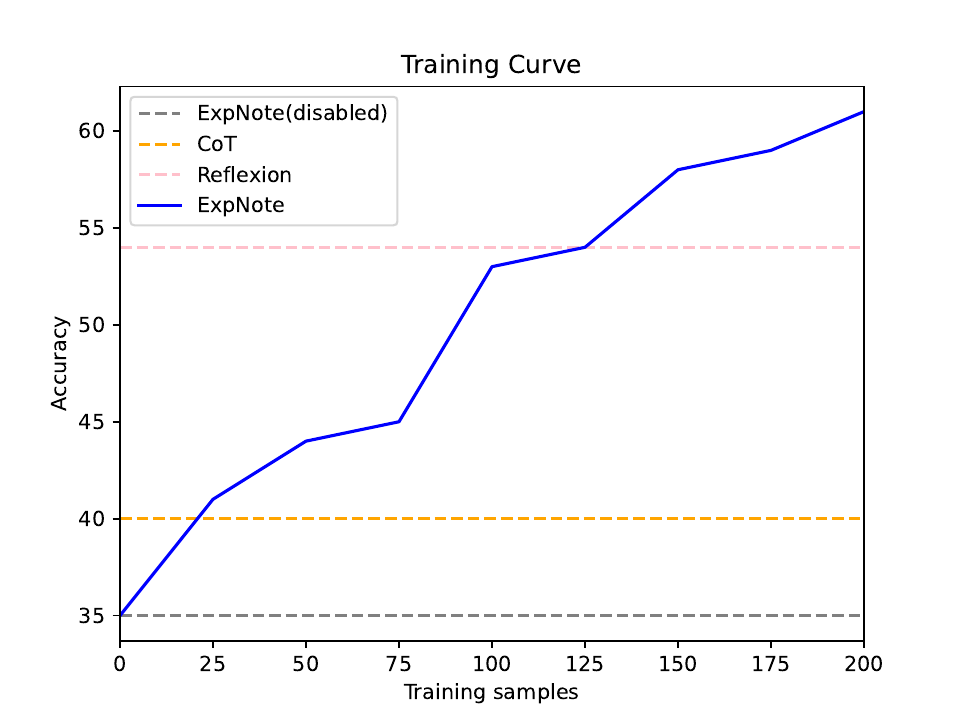}
    \end{adjustbox}
    \caption{The training curve in CLUTRR dataset.}
    \label{figure3}
\end{figure}

\subsection{Results}
As shown in Table \ref{table1}, the full ExpNote method achieved the best performance on all datasets, 20.5\% higher on average than the CoT method. TeachMe failed to outperform few-shot, as task-specific knowledge is hard to obtain without human feedback. Compared with Reflexion, note that even if we make Reflexion cheat to identify the failed trial with 100\% success rate, it still falls behind ExpNote.

Compared with other variants of ExpNote, \emph{disabled} retrieves no experience in the testing stage, thus degrading the performance of CoT (even worse) as expected. 
We also discovered that \emph{case} performs worse than full ExpNote although retrieving exactly the same cases for all of 4 tasks. 
We can then conclude that abstract knowledge or rules are more capable of helping LLMs to generalize to testing cases. 
Moreover, \emph{positive} and \emph{negative} both fall behind the full ExpNote while still outperforming baselines. 
We made an efficiency analysis in Appendix~\ref{effciency analysis} and the results show that experiences from both positive and negative cases are more efficient than the other respectively on two datasets. 
These results indicated that experiences learned from both positive cases and negative cases are useful for LLM to generalize to test sets. 

We also observed the performance changes of the model with the number of training samples. As shown in Figure~\ref{figure3}, in CLUTRR, ExpNote starts from training with 0 samples (equivalent to \emph{disabled}) and ends with training with 200 samples. The performance of ExpNote on the testing set continually grows with the number of training samples, showing that ExpNote continually learns new knowledge during the training stage.


\subsection{Improvement Analysis}
\label{improvement analysis}
\begin{figure}[t]
    \centering
    \begin{adjustbox}{max width=\columnwidth}
    \includegraphics[width=0.45\textwidth]{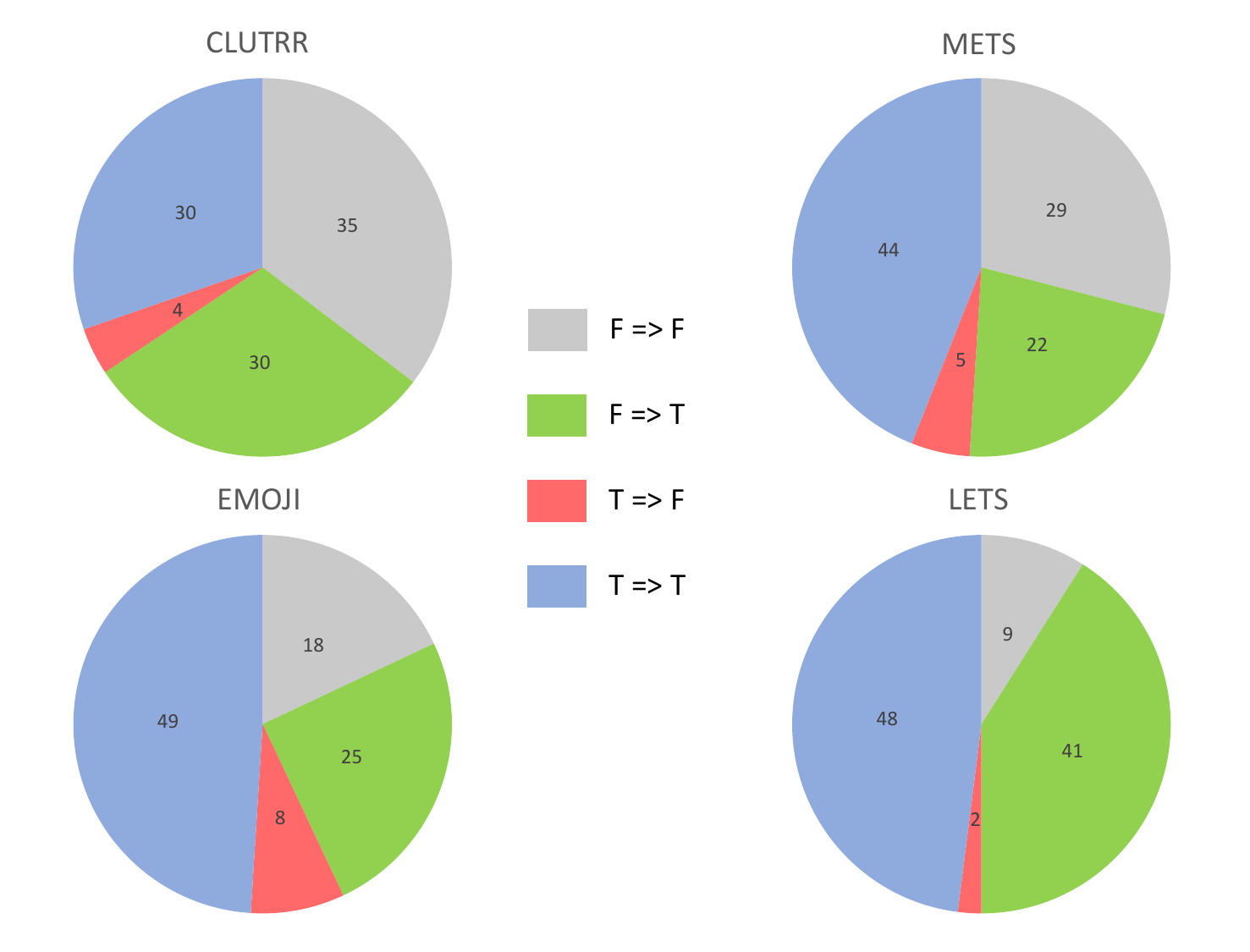}
    \end{adjustbox}
    \caption{Improvement analysis of ExpNote on 4 datasets.}
    \label{figure4}
\end{figure}

We also analyze how many cases are corrected by introducing experiences in each dataset. As shown in Figure~\ref{figure4}, we plot the distribution of cases in 4 conditions: 
\begin{itemize}[itemsep=1pt,topsep=1pt,parsep=0pt,leftmargin=*]
    \item{F => F: a case is originally answered incorrectly in \emph{disabled} and also answered incorrectly with ExpNote.}
    \item{F => T: a case is originally answered incorrectly in \emph{disabled} but answered correctly with ExpNote.}
    \item{T => T: a case is originally answered correctly in \emph{disabled} and also answered correctly with ExpNote.}
    \item{T => F: a case is originally answered correctly in \emph{disabled} but answered incorrectly with ExpNote.}
\end{itemize}

In Figure~\ref{figure4}, we demonstrate that ExpNote helps LLMs correct a certain amount of errors (the green part) at the cost of producing a few new errors (the red part) in all 4 datasets. And we can observe around 50\% incorrect answers in \emph{disabled} (gray + green) are corrected (green) with ExpNote. 

\section{Conclusion}
\noindent
In this paper, we propose ExpNote, an automated framework to help black-box LLMs adapt to specific downstream tasks by interacting with dynamic memory. We carried out experiments on multiple datasets from different tasks and showed that ExpNote can effectively improve the performance of LLMs better than other prompting methods.
We also found that the learned task-specific experiences help LLMs to better generalize than the original cases in the task, and experiences learned from both positive cases and negative cases are valuable.

\section*{Limitations}
Although ExpNote is able to empower the LLMs in various tasks, it may be less effective on these case-by-case tasks, like summarizing or creative writing. In these tasks, the cases share little common knowledge or rules, which makes it hard for ExpNote to help LLMs generalize.

\section*{Ethics Statement}
This paper proposes a method for augmenting black-box LLMs. All experiments are conducted on publicly available datasets. Thus there is no data privacy concern. Meanwhile, this paper does not involve human annotations, and there are no related ethical concerns.

\section*{Acknowledgements}
This work was supported by the National Key R\&D Program of China (2022ZD0160503) and the National Natural Science Foundation of China (No.62376270, No.61831022). This work was supported by the Strategic Priority Research Program of Chinese Academy of Sciences (No.XDA27020100),  Youth Innovation Promotion Association CAS and OPPO Research Fund.

\bibliography{anthology,custom}
\bibliographystyle{acl_natbib}

\appendix

\section{LETS Dataset}
\label{lets}
As existing symbolic reasoning datasets, such as word sorting in BIG-bench \cite{bigbench}, are designed to test the zero-shot reasoning ability of LLMs and always lack a training set, we therefore propose the LETS, a similar symbolic reasoning dataset while enabling LLMs to learn and generalize.

LETS require the language model to splice the letters at a given index of several words together. For example, given the query \emph{Splice the 5th letter of "sleep", the 2nd letter of "official", and the 5th letter of "neglect" together}, the model is supposed to output \emph{pfe} as the answer.

We randomly select 100 words with lengths of 4-10 as the vocabulary. To generate the training and testing set, for each instance, we randomly picked 3 different words from the vocabulary and randomly selected their indexes.

\section{Setup}
\label{setup}
For the LLM, we use ChatGPT (gpt-3.5-turbo) via Openai API calls.

For each task, due to the size limitations of the datasets themselves, we test all methods on 100 testing cases. In fact, a large amount of related work is also tested using samples of similar magnitude, such as TeachMe (OBQA, 500, \citealt{teachme}), ReAct (ALFWorld, 134; WebShop, 500, \citealt{react}), Reflexion (consistent with ReAct, \citealt{self-reflection}). 
Considering Expnote will interact with the environment multiple turns for a single case, the actual number of generations for LLMs can be 4 to 5 times higher.
And We adopt a minimal training set with it size 2:1 to the testing set (and 1:1 in EMOJI and LETS datasets). 

For all ExpNote variants, we write 2-3 ExpNote usage cases for the LLM as few-shot prompting; we choose $n=4$ for training (the LLM is able to take 4 extra actions to \textbf{THINK} and \textbf{NOTE} after obtaining the ground-truth of each case), and $n=0$ for testing (the LLM is not able to access to the ground-truth). 

For the retriever, we implemented a word-based retriever to retrieve experience by matching words in the query and the key of experience, and it retrieves up to $k=3$ experiences for each case in the testing stage.
When ExpNote fails to retrieve relevant experience, a failure prompt ``No relevant experience'' will be returned to the LLM.

\section{Effciency Analysis}
\label{effciency analysis}
We can define the efficiency of each type of experience as 
\begin{equation}
    \textrm{eff}(type) = \frac{\textrm{Perf}(type) - \textrm{Perf}(disabled)}{\textrm{Cnt}(type)}
\end{equation}

where $type$ refers to positive or negative, $\textrm{Perf}()$ represents the corresponding variant's performance in Table~\ref{table2}, and $\textrm{Cnt}(type)$ represents the number of experiences of that type. 
Then we can calculate the efficiency of positive and negative experiences based on the data in Table~\ref{table2}.

\begin{table}[t]
\centering
\begin{adjustbox}{width=\columnwidth}
\begin{tabular}{lcllll}
    \toprule
    type & CLUTRR & METS & EMOJI & LETS \\
    \midrule
    \emph{positive} & 0.219 & \textbf{0.042} & \textbf{0.643} & 0.341 \\
    \emph{negative} & \textbf{0.364} & 0.026 & 0.500 & \textbf{0.456} \\
    \bottomrule
\end{tabular}
\end{adjustbox}
\caption{Efficiency of Positive and Negative Experiences on 4 datasets.}
\label{table3}
\end{table}

As shown in Table~\ref{table3}, experiences from both positive and negative cases are more efficient than the other on two datasets. 
These results indicated that experiences learned from both positive cases and negative cases are useful for LLM to generalize to test sets. 




\section{Full Example}
\label{example}

\begin{figure*}[t]
    \centering
    \begin{adjustbox}{max width=\textwidth}    
    \includegraphics[width=1.8\textwidth]{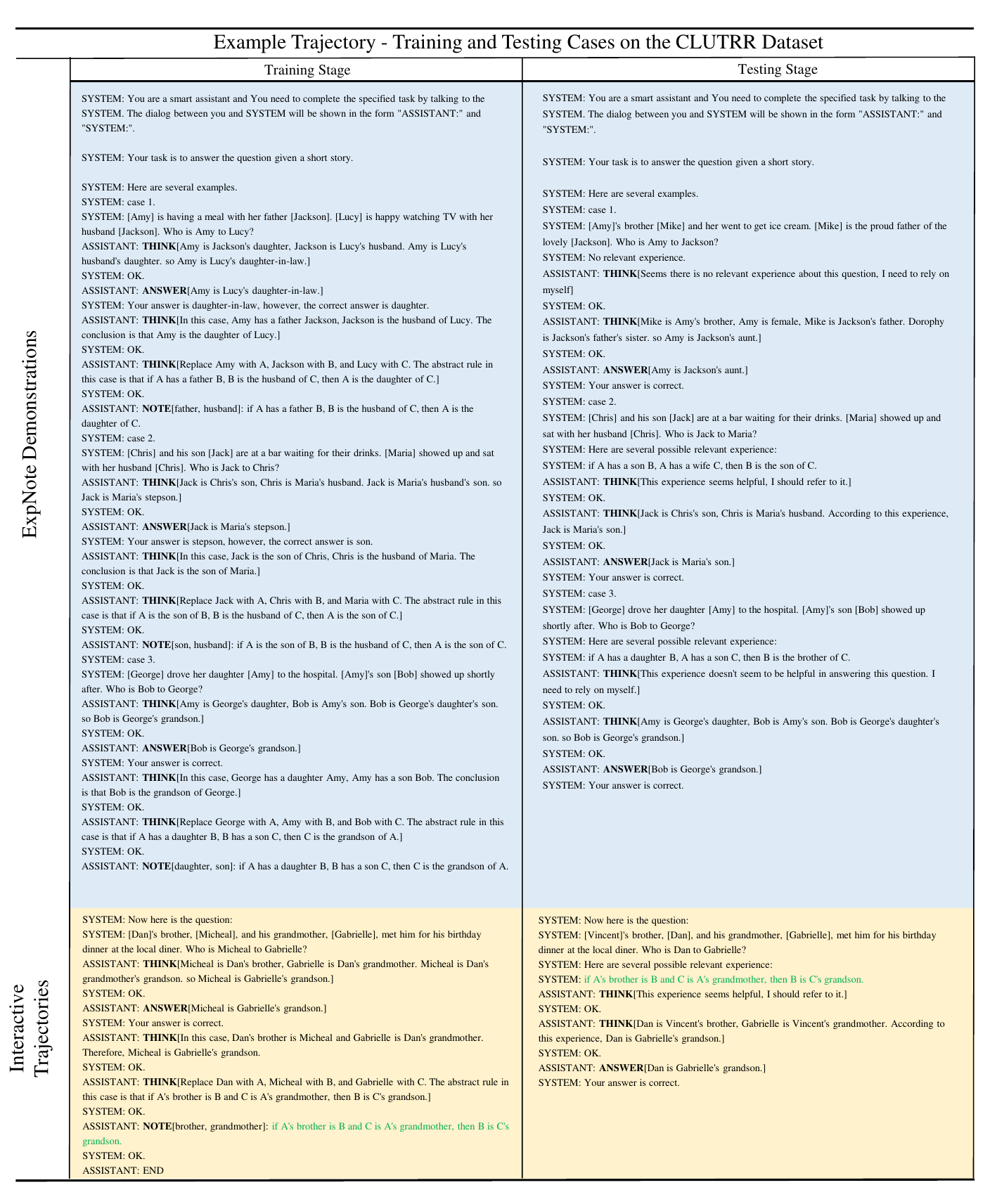}
    \end{adjustbox}
    \caption{The example trajectories of ExpNote on the CLUTRR Dataset. The left part is a training case while the right part is a corresponding testing case using that training experience. The blue parts are the ExpNote demonstrations (prompts $P_{train}$/$P_{test}$). The yellow parts are the interactive trajectories between LLM and ExpNote. The sentence highlighted in green is the learned experience.}
    \label{clutrr}
\end{figure*}

\begin{figure*}[t]
    \centering
    \begin{adjustbox}{max width=\textwidth}    
    \includegraphics[width=1.8\textwidth]{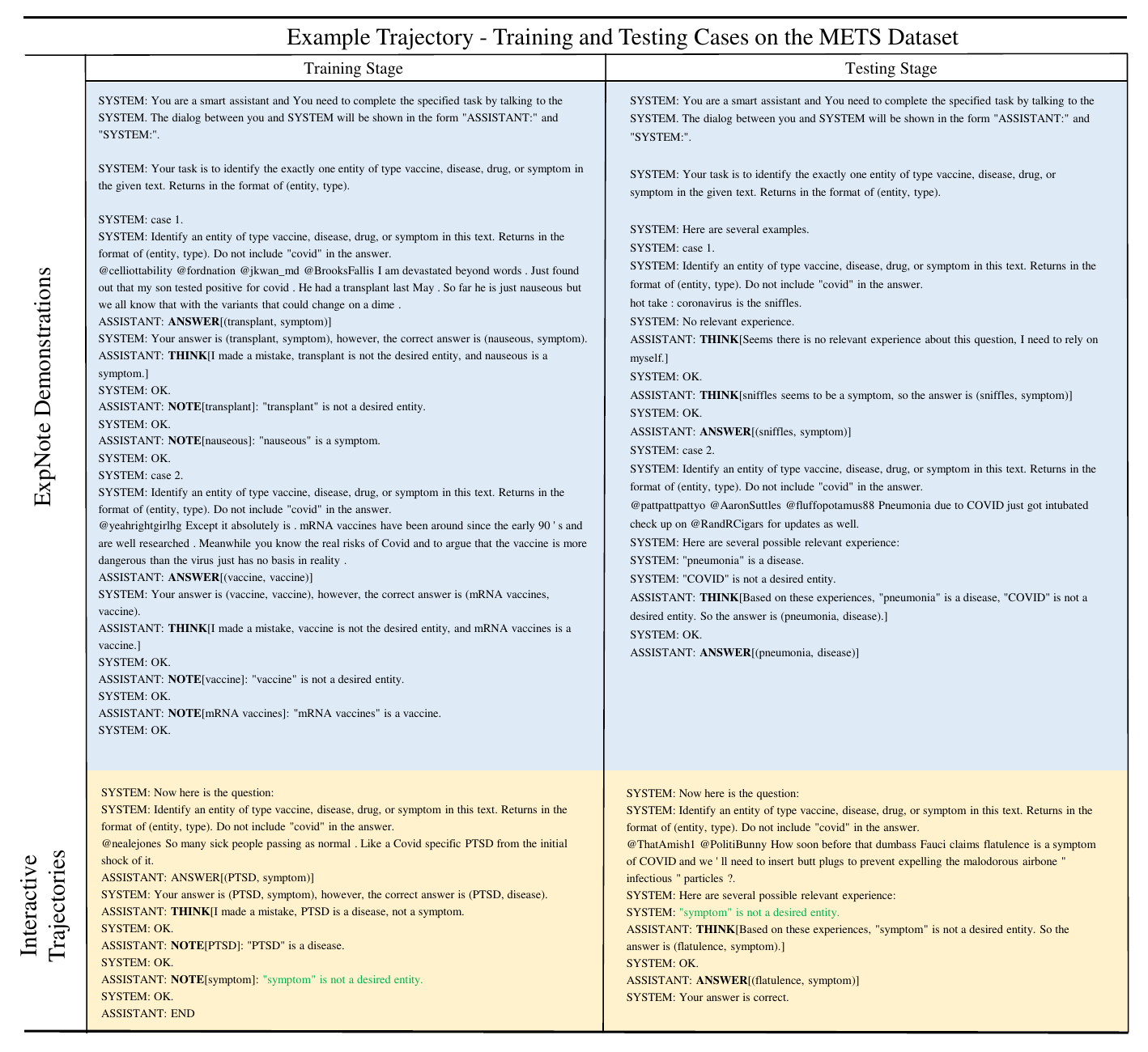}
    \end{adjustbox}
    \caption{The example trajectories of ExpNote on the METS Dataset. The left part is a training case while the right part is a corresponding testing case using that training experience. The blue parts are the ExpNote demonstrations (prompts $P_{train}$/$P_{test}$). The yellow parts are the interactive trajectories between LLM and ExpNote. The sentence highlighted in green is the learned experience.}
    \label{clutrr}
\end{figure*}

\begin{figure*}[t]
    \centering
    \begin{adjustbox}{max width=\textwidth}    
    \includegraphics[width=1.8\textwidth]{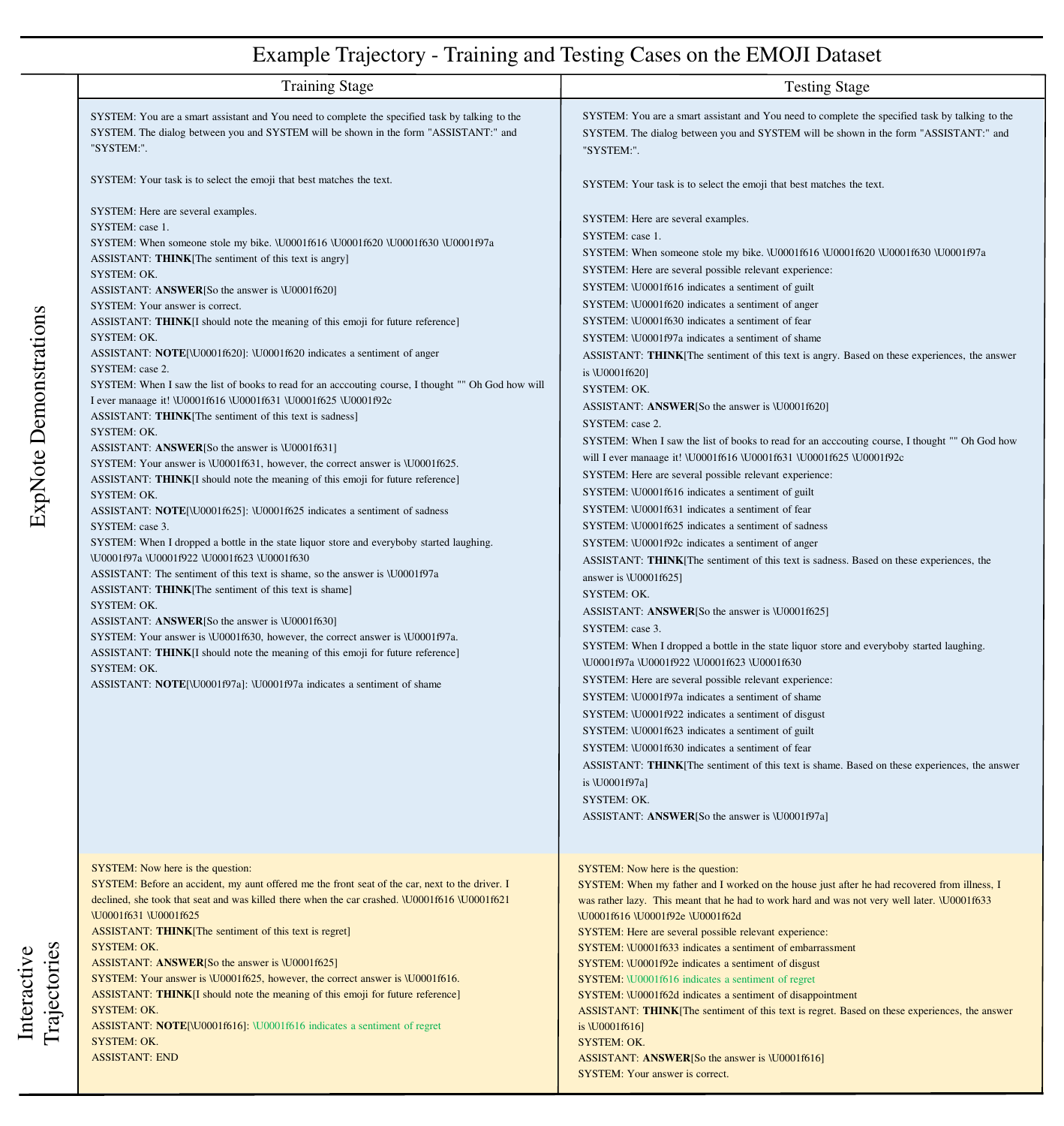}
    \end{adjustbox}
    \caption{The example trajectories of ExpNote on the EMOJI Dataset. The left part is a training case while the right part is a corresponding testing case using that training experience. The blue parts are the ExpNote demonstrations (prompts $P_{train}$/$P_{test}$). The yellow parts are the interactive trajectories between LLM and ExpNote. The sentence highlighted in green is the learned experience.}
    \label{clutrr}
\end{figure*}

\begin{figure*}[t]
    \centering
    \begin{adjustbox}{max width=\textwidth}    
    \includegraphics[width=1.8\textwidth]{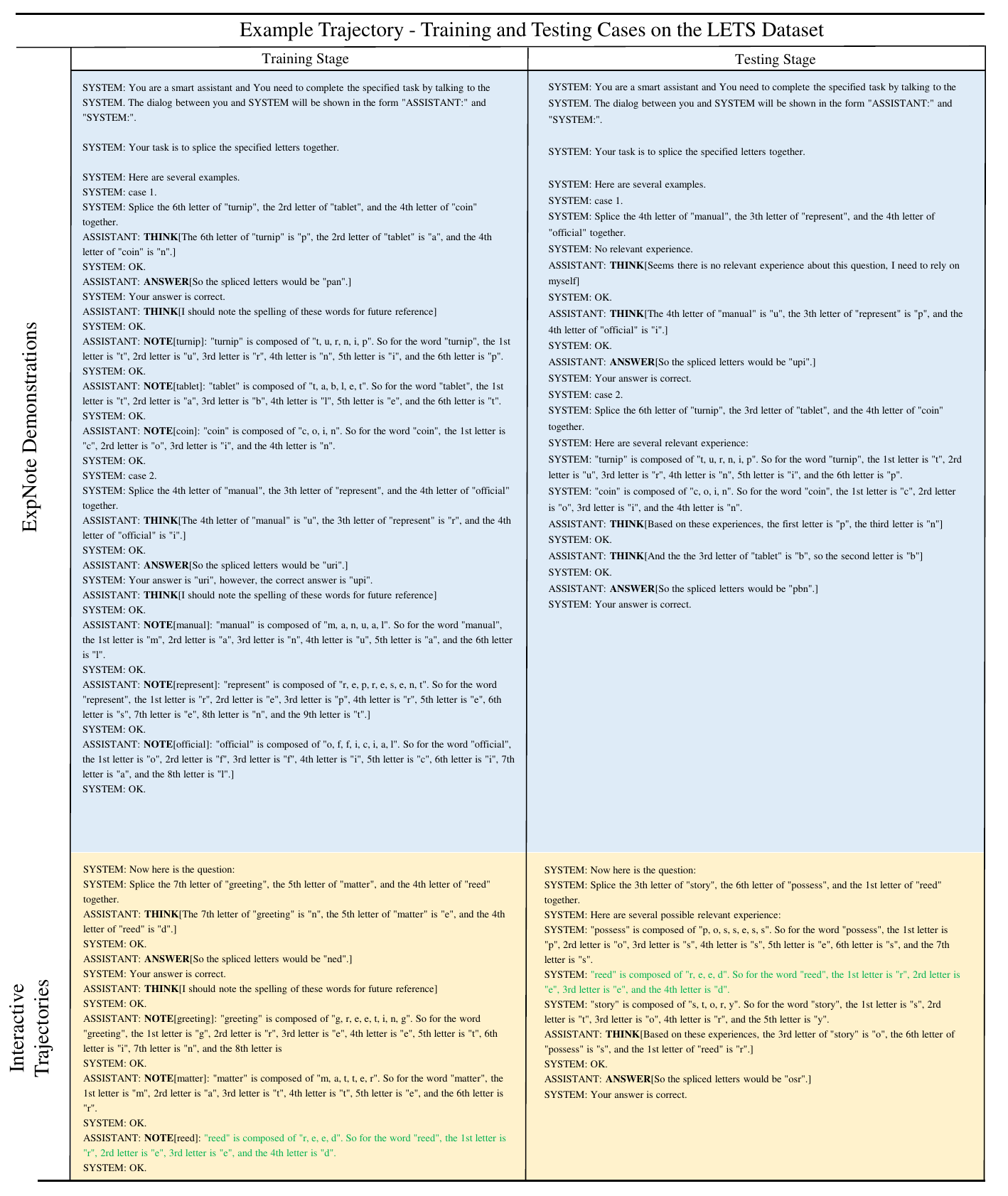}
    \end{adjustbox}
    \caption{The example trajectories of ExpNote on the LETS Dataset. The left part is a training case while the right part is a corresponding testing case using that training experience. The blue parts are the ExpNote demonstrations (prompts $P_{train}$/$P_{test}$). The yellow parts are the interactive trajectories between LLM and ExpNote. The sentence highlighted in green is the learned experience.}
    \label{clutrr}
\end{figure*}

\end{document}